\documentclass{article}
\usepackage{ijcai26}

\usepackage{times}
\usepackage{soul}
\usepackage{url}
\usepackage[hidelinks]{hyperref}
\usepackage[utf8]{inputenc}
\usepackage[small]{caption}
\usepackage{graphicx}
\usepackage{amsmath}
\usepackage{amsthm}
\usepackage{booktabs}
\usepackage{algpseudocode}
\usepackage[switch]{lineno}
\usepackage{algorithm2e}
\usepackage{amsfonts}

\usepackage{tikz}
\usetikzlibrary{arrows.meta, positioning}

\tikzset{
  block/.style={
    rectangle, draw, rounded corners, minimum height=1cm, minimum width=2cm, align=center
  }
}

\theoremstyle{definition}
\newtheorem{definition}{Definition}

\title{
  The Law of Task-Achieving Body Motion: Axiomatizing Success of Robot Manipulation Actions
}

\author{
Malte Huerkamp$^1$
\and
Jonas Dech$^1$\And
Michael Beetz$^{1}$
\affiliations
$^1$AICOR Institute for Artificial Intelligence\\
University of Bremen\\
Bremen, Germany\\
\emails
\{huerkamp, jdech, mbeetz\}@uni-bremen.de
}

\begin{document}

\maketitle

\begin{abstract}
Autonomous agents that perform everyday manipulation actions need to ensure that their body motions are semantically correct with respect to a task request, causally effective within their environment, and feasible for their embodiment.
In order to enable robots to verify these properties, we introduce the Law of Task-Achieving Body Motion as an axiomatic correctness specification for body motions.
To that end we introduce scoped Task-Environment-Embodiment (TEE) classes that represent world states as Semantic Digital Twins (SDTs) and define applicable physics models to decompose task achievement into three predicates: SatisfiesRequest for semantic request satisfaction over SDT state evolution; Causes for causal sufficiency under the scoped physics model; and CanPerform for safety and feasibility verification at the embodiment level.
This decomposition yields a reusable, implementation-independent interface that supports motion synthesis and the verification of given body motions. It also supports typed failure diagnosis (semantic, causal, embodiment and out-of-scope), feasibility across robots and environments, and counterfactual reasoning about robot body motions.
We demonstrate the usability of the law in practice by instantiating it for articulated container manipulation in kitchen environments on three contrasting mobile manipulation platforms.
\end{abstract}

\section{Introduction}

Everyday robot manipulation such as opening a drawer, moving objects or pouring substances has to succeed in human environments where objects, mechanisms, and physical conditions vary to much to be completely specified at design time. Learning-based control and robot foundation models~\cite{rt1} have improved robustness to perception noise and situational variation, but cannot guarantee correctness. These systems can reliably apply forces while acting on the wrong joint, violating intent, or producing unsafe side effects. For example, an agent can try to ``open the drawer'' by pulling the handle, but fail because the drawer is latched, or succeed by applying excessive force that risks damage. What is missing is a systematic account of why a proposed motion should be trusted as a task solution in the physical world and how failures can be identified.

To approach this we follow a pattern from algorithm theory where for hard computational problems, not a single method that should be correct and efficient on all inputs is searched for. Instead, scoped problem classes with explicit structure and assumptions are defined and then algorithms that are correct within the class are designed and optimized by exploiting the defined structure. We argue that everyday manipulation should be treated in the same way. Then correctness and guarantees can be stated relative to a declared scope, and the structure of that scope should be made explicit and exploited in algorithm design.

\begin{figure}
  \centering
  \includegraphics[width=\linewidth]{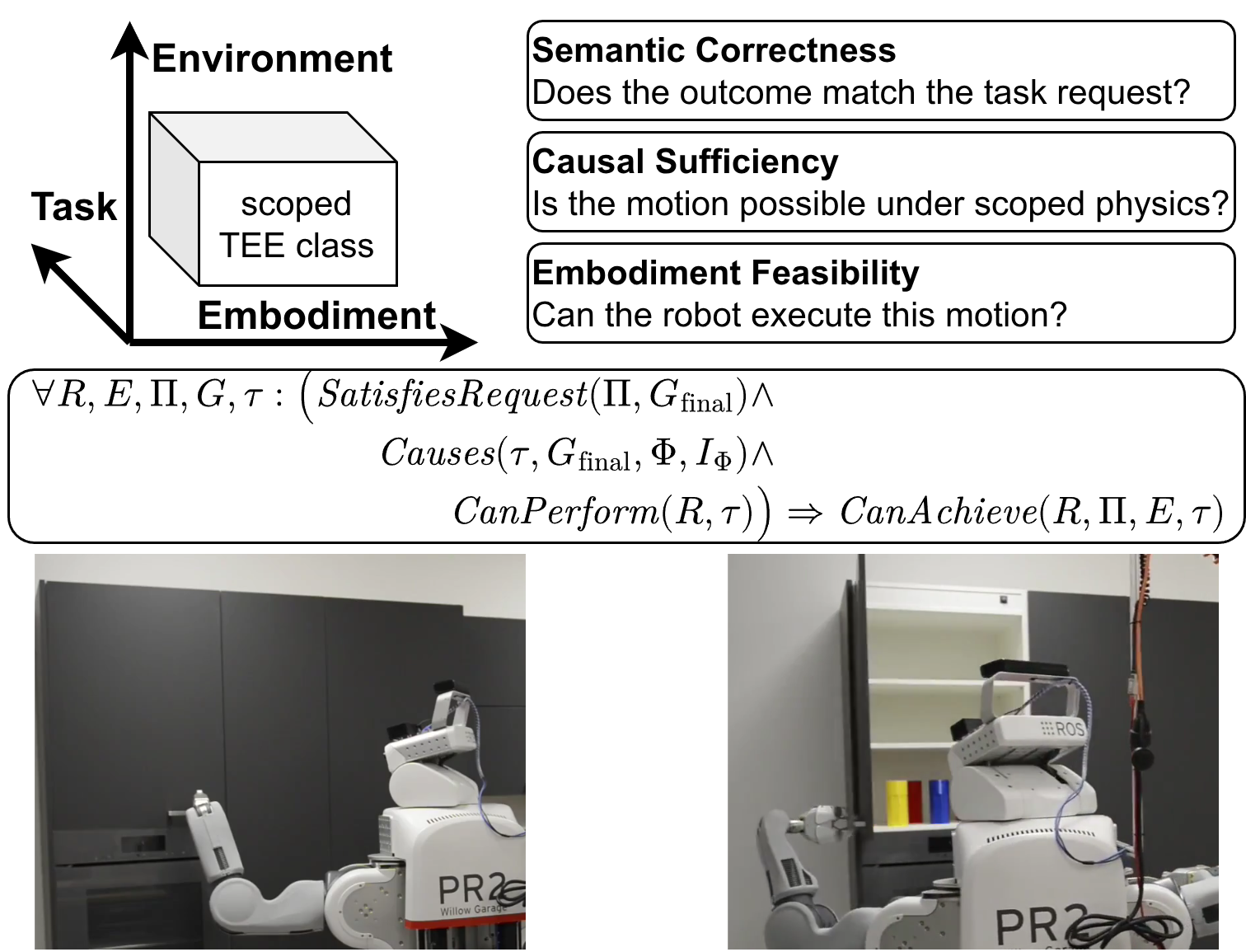}
  \caption{The Law of Task Achieving Body Motion(middle box) axiomatizes the success of robot manipulation actions under a scoped physics model (top left box) and by splitting success validation into the dimensions of semantic correctness, causal sufficiency and embodiment feasibility (top right box).}
  \label{fig:va}
\end{figure}

Therefore, we advocate the use o assertion-enabled reasoning for everyday manipulation. Its purpose is to scope a family of Task--Environment--Embodiment (TEE) combinations and to axiomatize within that scope what counts as success, what interactions can produce which changes, and what the robot can realize, a visual summary of that can be seen in Fig.~\ref{fig:va}. These axioms provide usable facts that define truth conditions an agent can check and compose. The resulting guarantees are conditional to its designed scope and the right behavior outside that scope is defined as principled abstention.

Concretely, within a fixed scoped class, task achievement can be decomposed into three assertions. First, a predicted change in the world state satisfies the task request. Second, there exists an interaction in the world consistent with the scoped physics that is sufficient to produce that change. Third, the robot can generate a body motion that realizes such an interaction. If all three hold jointly, then executing the corresponding body motion achieves the task within the declared scope. This decomposition is deliberately generator-agnostic, as the candidate motion may come from planning, control, learning, or LLM-based proposal mechanisms. Correctness is then assessed against the scoped assertions rather than the generator's internal representations. 
The goal is a single axiomatic criterion that combines intention, physical limitations and embodiment feasibility into a single truth-based representation

Our research hypothesis is that for practically relevant manipulation domains we can state an axiomatic characterization of task-achieving body motions and use predicate-level reasoning to certify and diagnose candidate motions, while also supporting synthesis by searching for motions that satisfy the axioms within scope. To make the domains of quantification explicit, we represent world state evolution with semantic digital twins (SDTs). They are structured scene-graph states whose entities, relations, and physical attributes support both semantic predicates and physics/capability predicates. Scoping is enforced through validity regimes for physics parameters.

In this paper we instantiate this framework for articulated container manipulation in kitchen environments using multiple robot embodiments. 
In summary, we contribute: (1) the Law of Task-Achieving Body Motion as an axiomatic correctness specification for manipulation, (2) assertion-enabled scoping for TEE/physics classes with principled abstention, (3) typed diagnosis and meta-reasoning derived from predicate outcomes, and (4) executable SDT-based instantiations with a reproducible experimental setup.

The rest of this paper is structured as follows: In Section~\ref{sec:formalization} we define the conceptual groundwork, followed by Section~\ref{sec:law} were we present the formalization of the law of task-achieving body motion. We evaluate the presented framework in Section~\ref{sec:evaluation}, discuss related work in Section~\ref{sec:rel} and finally conclude the paper in Section~\ref{sec:conc}.

\section{Conceptualization and Formalization}
\label{sec:formalization}

We require a formal substrate that (i) represents the causal structure of the world in a way that supports logical quantification, (ii) defines correctness over motion trajectories rather than only terminal states, and (iii) binds the validity of such claims through systematic scoping. This section introduces that substrate: the Body Motion Problem (BMP), Task--Environment--Embodiment (TEE) classes, Semantic Digital Twins (SDTs), and a factored physics model that separates robot body motion from environment physics.

\subsection{The Body Motion Problem}
\label{subsec:bmp}

We employ a trajectory-centric view of manipulation in which success depends on the temporal evolution of contacts, flows, and geometric relations.

Let $\mathcal{S}$ be the state space and $\mathcal{A}$ the action space. A trajectory $\tau$ over horizon $H$ is a sequence of state--action pairs
\begin{equation}
\tau = \bigl( (s_0,a_0),(s_1,a_1),\dots,(s_H,a_H) \bigr) \in (\mathcal{S} \times \mathcal{A})^{H+1}.
\end{equation}

Rather than maximizing a scalar reward, we formulate the BMP as a constraint satisfaction and optimization problem over trajectory families.

\begin{definition}[Manipulation Task Specification]
A manipulation task specification is a tuple $\Pi = \langle \Pi_{\text{goal}}, \Pi_{\text{avoid}}, J \rangle$, where
\begin{itemize}
\item $\Pi_{\text{goal}} \subseteq (\mathcal{S} \times \mathcal{A})^{\ast}$ is the set of semantically valid trajectories,
\item $\Pi_{\text{avoid}} \subseteq (\mathcal{S} \times \mathcal{A})^{\ast}$ is the set of forbidden trajectories,
\item $J : (\mathcal{S} \times \mathcal{A})^{\ast} \rightarrow \mathbb{R}$ is a quality functional (e.g., energy, time).
\end{itemize}
A trajectory $\tau$ is a valid solution iff $\tau \in \Pi_{\text{goal}}$ and $\tau \cap \Pi_{\text{avoid}} = \emptyset$.
\end{definition}

The BMP separates correctness (membership in $\Pi_{\text{goal}} \setminus \Pi_{\text{avoid}}$) from quality defined by the value of $J$ and defines correctness over entire body-motion trajectories, which is the level at which the Law of Task-Achieving Body Motion will operate.

\subsection{Task--Environment--Embodiment Classes and Conservative Scoping}
\label{subsec:scope}

To obtain formal guarantees in open-world manipulation, we restrict attention to subspaces of tasks, environments, embodiments and their associated physics. Within such a scope, BMP instances are well-defined and share common modeling assumptions.

\begin{definition}[Task--Environment--Embodiment Class]
A manipulation scope, or Task--Environment--Embodiment Class, is a tuple
\begin{equation}
D = \langle \mathcal{T}, \mathcal{E}, \mathcal{R}, \Phi, I_{\Phi} \rangle,
\end{equation}
where
\begin{itemize}
\item $\mathcal{T}$ is a family of task types, 
\item $\mathcal{E}$ is a family of environment models, 
\item $\mathcal{R}$ is a familiy of robot models, 
\item $\Phi$ is the governing physics model, 
\item $I_{\Phi}$ is a set of validity intervals for physical parameters 
\end{itemize}
\end{definition}

A TEE class $D$ specifies that we consider tasks $\mathcal{T}$ in environments $\mathcal{E}$ for embodiment's $\mathcal{R}$ under physics $\Phi$ that is trusted only for parameter values in $I_{\Phi}$. Concrete BMP instances live inside a chosen TEE class and inherit its modeling assumptions.

The set $I_{\Phi}$ defines the domain within which the analytical and simulation models associated with $\Phi$ are trusted. If a real-world estimate $p$ of the relevant parameters satisfies $p \notin I_{\Phi}$, the framework abstains.
This conservative interval scoping strategy allows to state guarantees relative to $D$.

\subsection{Semantic Digital Twins as Structured State Representation}
\label{subsec:sdt}

To quantify over classes of objects (e.g., ``all containers'') and express task conditions as logical predicates, we replace unstructured state vectors $s \in \mathcal{S}$ with a structured SDT, which serves as the semantic state space for TEE classes.

\begin{definition}[Semantic Digital Twin]
An SDT state is a directed, attributed graph $G = (V, E_G, P)$, where
\begin{itemize}
\item $V$ is a set of named entities, 
\item $E_G$ is a set of spatial and semantic relations between entities, 
\item $P : V \rightarrow \mathcal{B}_{\text{prop}}$ maps each entity to a record of physical and geometric properties required by $\Phi$ 
\end{itemize}
\end{definition}
 
For each $G$ and action $a$, the map $P$ contains all data required to instantiate a physics simulation $\mathrm{Sim}(G,a)$ under $\Phi$. This allows us to lift the motion task specification $\Pi$ from abstract state sequences to scene-graph sequences $G_0,G_1,\dots$ and to treat $\Pi_{\text{goal}}$ and $\Pi_{\text{avoid}}$ as sets of SDT trajectories.

The SDT thus provides the explicit naming and structural access that the Law will require. Its predicates will be formulated as logic-based queries against $G$ and its evolution, and TEE classes will specify which SDT schemas and physics models are admissible.

\subsection{Decomposed Causality}
\label{subsec:decomposed-causality}

We explicitly separate robot agency from environmental causality by factoring the state transition dynamics within each TEE class, using its SDT schema and physics model.

Let $G_{\text{robot}}$ and $G_{\text{env}}$ be the subgraphs corresponding to the robot and its environment, respectively. The evolution of the world is governed by
\begin{equation}
G_{t+1} = \mathrm{Sim}(G_t, a_t) \equiv \Phi \bigl(G_{\text{env},t}, \Gamma(G_{\text{robot},t}, a_t)\bigr),
\end{equation}
where
$\Gamma$ is a body motion function that describes how the robot state evolves when performing action $a_t$.

This decomposition isolates
robustness, which is the stability of $\Gamma$ under noise and perturbations from
correctness, which is the success of $\Phi$ in producing trajectories in $\Pi_{\text{goal}} \setminus \Pi_{\text{avoid}}$.

Therefore, BMP instances specify what counts as a correct trajectory, TEE classes specify where and under which physical assumptions we reason, SDTs provide the structured state space in which tasks and dynamics are expressed, and the factored dynamics separate body motions from environmental responses. 
The Law of Task-Achieving Body Motion, introduced in Section~\ref{sec:law}, will relate task specifications $\Pi$, SDT trajectories $G_0,\dots,G_H$, and body motions induced by $\Gamma$ under a given TEE class $D$, and will define when a candidate motion is certified as task-achieving, causally sufficient, and embodiment-feasible within that scope.

\section{The Law of Task--Achieving Body Motion}
\label{sec:law}

Given the Body Motion Problem, TEE classes, and SDTs, we now introduce the Law of Task-Achieving Body Motion as the core contribution of this work. This universally quantified axiom schema serves as a logical gatekeeper between high-level task semantics and low-level physical execution. 
It states that a body motion is a valid solution to a manipulation task in a given TEE class iff three conditions jointly hold: (i) the outcome satisfies the task request, (ii) the motion is causally sufficient to produce that outcome under the scoped physics, and (iii) the motion is feasible for the specific robot embodiment.

\subsection{Axiom Schema over TEE Classes}
\label{subsec:law-schema}

Let $D = \langle \mathcal{T}, \mathcal{E}, \mathcal{R}, \Phi, I_{\Phi} \rangle$ be a TEE class as defined in Section~\ref{subsec:scope}. Let $R \in \mathcal{R}$ be a robot, $E \in \mathcal{E}$ an environment state represented as an SDT scene graph $G$, $\Pi \in \mathcal{T}$ a task specification, and $\tau$ a candidate motion trajectory that, under the dynamics of $D$, induces a final scene graph state $G_{\text{final}}$. The Law is given by:
\begin{equation}
\begin{aligned}
\forall R, E, \Pi, G, \tau : 
\bigl(
  \mathit{SatisfiesRequest}(\Pi, G_{\text{final}}) \wedge & \\
  \mathit{Causes}(\tau, G_{\text{final}}, \Phi, I_{\Phi}) \wedge 
  \mathit{CanPerform}(R, \tau) 
\bigr) & \\
\Rightarrow
\mathit{CanAchieve}(R, E, \Pi, \tau)
\end{aligned}
\label{eq:law}
\end{equation}
Here, $\mathit{CanAchieve}(R, E, \Pi, \tau)$ asserts that, within TEE class $D$, robot $R$ can successfully and safely execute task $\Pi$ in environment $E$ using motion $\tau$. 
The law does not specify the implementation of the predicates, but it definees their semantics.

\subsection{Predicate 1: Semantic Correctness}
\label{subsec:predicate-semantic}

The first predicate,
\[
\mathit{SatisfiesRequest}(\Pi, G_{\text{final}}),
\]
checks that a final SDT graph state matches the intent of the task specification. Let $\Pi_{\text{goal}}$ denote the goal component of $\Pi$, expressed as a logical query over SDT graphs. Then we can define
\begin{equation}
\mathit{SatisfiesRequest}(\Pi, G_{\text{final}})
\;\;\Leftrightarrow\;\;
G_{\text{final}} \models \Pi_{\text{goal}}.
\label{eq:satisfies}
\end{equation}

This predicate can be implemented via structured queries over scene-graph representations~\cite{rosinol_kimera,qsr_3d} or by integrating knowledge bases for semantic reasoning~\cite{knowrob}. 

\subsection{Predicate 2: Causal Sufficiency under Scoped Physics}
\label{subsec:predicate-causal}

The second predicate,
\[
\mathit{Causes}(\tau, G_{\text{final}}, \Phi, I_{\Phi}),
\]
asserts that within the TEE class $D$, the motion $\tau$ is a physically valid explanation of the transition from the initial SDT state to $G_{\text{final}}$ under the physics model $\Phi$. It operationalizes conservative interval scoping by requiring both that the relevant physical parameters are in scope and that simulation agrees with the predicted outcome:
\begin{equation}
\begin{aligned}
\mathit{Causes}(\tau, G_{\text{final}}, \Phi, I_{\Phi})
\;\;\Leftrightarrow\;\;
\bigl( \mathit{Params}(G) \in I_{\Phi} \bigr)
\wedge \\
\bigl( \Phi_{\text{sim}}(G, \tau) \approx G_{\text{final}} \bigr).
\end{aligned}
\label{eq:causes}
\end{equation}

In practice, $\Phi_{\text{sim}}$ is instantiated using physics engines such as MuJoCo~\cite{mujoco}, Drake~\cite{drake}, or PyBullet~\cite{pybullet}, which simulate rigid-body dynamics, contact mechanics, and articulation constraints. Differentiable physics approaches~\cite{degrave_diffphys} enable gradient-based verification, while digital twin methodologies~\cite{glaessgen_digital_twin} can ensure model fidelity through continuous validation against real-world data.

If $\mathit{Params}(G) \notin I_{\Phi}$, the predicate fails structurally and the system reports an \textsc{OutOfScope} condition. If parameters are in scope but simulation does not reproduce $G_{\text{final}}$, the predicate fails for causal reasons.

\subsection{Predicate 3: Embodiment Feasibility}
\label{subsec:predicate-embodiment}

The third predicate,
\[
\mathit{CanPerform}(R, \tau),
\]
checks whether the candidate motion $\tau$ is executable by robot $R$, independently of task success. Let $K_R$ denote the kinematic and dynamic limits of $R$. Writing $\tau$ in joint space as joint position, velocity and accelaration $(q_t, \dot{q}_t, \ddot{q}_t)$ over time, we can define:
\begin{equation}
\begin{aligned}
\mathit{CanPerform}(R, \tau)
\;\;\Leftrightarrow\;\;
\forall t \in \tau :
\bigl( (q_t, \dot{q}_t, \ddot{q}_t) \in K_R \wedge \\
\neg \mathit{SelfCollision}(q_t) \bigr).
\end{aligned}
\label{eq:canperform}
\end{equation}

This predicate can be implemented using inverse kinematics solvers such as TRAC-IK~\cite{tracik} or the MoveIt! framework~\cite{moveit}, which incorporate joint-limit constraints and redundancy resolution. Whole-body control methods~\cite{whole_body_mpc} extend feasibility checking to include dynamic constraints, stability criteria, and prioritized task execution. Collision detection is handled by geometric libraries such as FCL~\cite{fcl_collision}, which provide efficient proximity queries for self-collision and environment-collision checking.

\subsection{Multi Use Property}
\label{subsec:relational-use}

As the Law is formulated as predicate logic relations over $(R, E, \Pi, G, \tau)$, it can be used beyond one-way verification. With the predicates implemented as logical relations over SDT structures, different inference modes are possible by choosing which arguments and predicate outcomes are treated as known or unknown.

\textbf{Motion generation.} Given fixed instances of $(R, E, \Pi, G)$ search for a motion $\tau$ such that all three predicates hold and therefore $\mathit{CanAchieve}(R,E,\Pi,\tau)$ is true. In this mode, the Law acts as a task-level specification for underlying motion generators. We discuss this functionality in more detail in usage mode 1 of Section \ref{sec:evaluation}.

\textbf{Explanation and diagnosis.} Given an observed $\tau$ and the fixed context $(R, E, \Pi, G)$ a multitude of diagnostics capabilities are possible.
The evaluation of $\mathit{CanPerform}(R, \tau)$ checks whether the motion is embodiment feasible, while the evaluation of the other two predicates can make statements about the causal sufficiency and semantic correctness of the motion.
Evaluating all three predicates together then verifies the success of the motion in the defined scope.
We discuss this in more detail in the usage modes 2 and 3 presented in Section \ref{sec:evaluation}.

\textbf{Counterfactual reasoning.} 
Counterfactual reasoning~\cite{2001_counterfactual} is the process of finding answers to 'What If?' question.
We can realize that by employing any of the previous inference modes but deliberately changing instances of $(R, E, \Pi, G, \tau)$ before predicate evaluation to create alternative scenarios.
For example, asking which other robots would achieve $\Pi$ in the same SDT, or which environment changes make a previously impossible task satisfiable.

All of these inference modes share the same axiomatic core and differ only in which instances of $(R, E, \Pi, G, \tau)$ and which predicate outcomes are treated as given or to be solved for. This relational perspective makes the Law a reusable logical interface for synthesis, verification, diagnosis, feasibility analysis, and counterfactual design across TEE classes.

\section{Evaluation}
\label{sec:evaluation}
In this section we instantiate the Law for the use case of articulated container manipulation in two kitchen environments shown in Fig.\ref{fig:causesResults}.
On the basis of this example we will discuss four usage modes of the law in practice.
First, the generation of robot body motions from a task request, second inferring effects and suitable task request from a motion candidate, third employing the law as a framework for learning from observed motions and third counterfactual reasoning by evaluating the law over different task, environment and embodiment combinations.
Before doing that we explain the assumptions for the scoped TEE classes and the predicate implementation in this use case.
For the usage modes 1, 2 and 4 we provide implementations in a web-based application that can be made available after acceptance, as the code includes dependencies that would undermine the double blind review process.

\subsection{Articulated Container Manipulation Setup}
\label{subsec:experimental-setup}

We define the TEE class $D_{\mathrm{artic}}$ to represent the domain of opening and closing containers (cupboard doors, drawers and dishwasher and oven doors) in kitchen environments using mobile manipulators. It consists of:

\paragraph{Task class.}
A Task class contains the set of all trajectories (according to Definition 1) that satisfy a given task request. 
The task set $\mathcal{T}_{\mathrm{artic}}$ comprises tasks $\textit{open}(c)$ and $\textit{close}(c)$ for each container $c$. When the reference to a container is omitted the meaning of the tasks becomes to open/close any container in the environment. 

\paragraph{Environment class.}
The environment family $\mathcal{E}_{\mathrm{artic}}$ contains kitchen scenes, each represented as an SDT.
Besides the articulated kinematic model of the kitchens, the SDT representation contains semantic annotations for the entities \textit{Body}, \textit{Door}, \textit{Drawer}, \textit{Handle}, \textit{Joint}, \textit{ActiveJoint}, \textit{PassiveJoint}, \textit{Container}.
These annotations can be inferred automatically for new kitchen models, when an URDF file with appropriately named link names is provided.
Based on these annotations the relations: $\textit{HasHandle}(c,h)$, $\textit{HasJoint}(c,j)$, $\textit{HasBody(c,b)}$ are defined.
These can also be used as queries to the SDT in a predicate logic format.
The physical properties, like pose, collision mesh, mass and joint limits of each body $b$ or entity $v$ are mapped by $P(v)$.
To conclude the environment class definition for $D_{\mathrm{artic}}$ semantic annotations are defined over articulation state variables, i.e. the \textit{ActiveJoint}. For a container $c$ with their active joint value characterized by $p_c$, their open and closed state is defined as:
\begin{equation}
\begin{aligned}
  \textit{open}(c) \equiv p_c \ge 0.9 p_{\text{c}}^{\text{upper-limit}}, \\
  \textit{closed}(c) \equiv p_c \le p_c^{\text{lower-limit}} + \epsilon,
\end{aligned}  
  \label{eq:semantic-predicates}
\end{equation}
where $\epsilon$ is an threshold value.

\paragraph{Embodiment class.}
We define three robot platforms with distinct kinematic and morphological characteristics in $D_{\mathrm{artic}}$.
The \textbf{PR2} is a larger humanoid mobile robot with two 7-DoF arms.
The \textbf{TIAGo++} is similar to the PR2 but smaller, therefore it needs less space to navigate but cannot reach as high as the PR2.
The third robot is the \textbf{Stretch} from Hello Robot. It has a small base footprint and one arm consisting of only prismatic joints. It is less flexible than the other robots but can move its arm lower than them.
These differences in base footprint, reach envelope, and kinematic structure should impact which containers each robot can manipulate, making embodiment feasibility checking critical in the experiments presented below.
For this use case we consider the kinematics models of the robots.
They are part of the SDT for the environment.
Relevant semantic annotations are the available gripper and their tool frames.
Physical properties we consider here are the joint limits for position and velocity, as well as collision meshes.

\paragraph{Physics model and state space.}
Finally, we define the physics model $\Phi_{\text{artic}}$ and its scope $I_{\Phi \text{artic}}$ for $D_{\mathrm{artic}}$.
For the sake of clarity in the presentation of the following use case, we limit $\Phi_{\text{artic}}$ to rigid body kinematics of articulated mechanisms. Future work will include domains with more realistic physics models, by utilizing appropriate general purpose physics simulators.
The definition of $I_{\Phi \text{artic}}$ therefore reduces to the position, velocity and acceleration joint limits of the environment and robot models. 

\paragraph{Predicate implementations.}
For the experiment we implemented the three predicates of the Law as follows:
\begin{itemize}
\item $\mathit{SatisfiesRequest}(\Pi_{\text{artic}}, G_{\text{final}})$: Queries the SDT to select the semantic states defined in Equation \ref{eq:semantic-predicates} that fit the task request $\Pi_{\text{artic}} \in \mathcal{T}_{\text{artic}}$. As this is similar to a lookup table, it also works backwards to select suitable task requests when a final state is known.
\item $\mathit{Causes}(\tau, G_{\text{final}}, \Phi_{\text{artic}}, I_{\Phi \text{artic}})$: Replays motion $\tau$ through $\Phi_{\text{artic}}$ to verify that the evolution entails the desired semantic state change while respecting articulation constraints. When the motion is unknown it generates $\tau$ using the forward-kinematics expressions of the environment models in $\Phi_{\text{artic}}$.
We implemented this with a method inspired by~\cite{roefer2022kineverse}.
Motions can be represented either as one dimensional trajectories of the articulation variable or as trajectories of any other point in Cartesian space that is connected to that articulation variable, e.g. the handle of a container.
Both representations are equivalent because $\Phi_{\text{artic}}$ includes forward-kinematics expressions of the environment models.
\item $\mathit{CanPerform}(R, \tau)$: Parameterizes a constraint and optimization based motion planning method~\cite{stelter25giskard} with $\tau$ as a reference trajectory for one gripper of the robot $R$. It calculates a motion for each gripper of the robot until a feasible motion is found. It fails if there is no feasible motion for all grippers.
For the motion planning the robot is placed at a fixed position in the world and rotated to look towards the goal. Then the motion planner is tasked to calculate a whole-body motion that first approaches the handle of a container with the gripper and then follows $\tau$ with the gripper, where $\tau$ was transformed into a motion of the handle using forward-kinematics expressions of the environment models.  
\end{itemize}

With respect to these definitions of $D_{\mathrm{artic}}$ and the predicates, we can state  that motions satisfying all three predicates are semantically and causally correct as well as embodiment feasible.

\subsection{Different Usage Modes}
\label{subsec:multi-use-demonstration}

We demonstrate how the Law can be used to axiomatize the success of manipulation actions in $D_{\mathrm{artic}}$ by investigating how it enables motion generation, verification, and feasibility analysis via counterfactual reasoning.
We structure this by looking at four different usage modes of the Law in the context of the container opening domain.
\paragraph{Usage Mode 1: Motion Generation from Task Specifications.}

\begin{figure}
  \centering
  \includegraphics[width=0.9\linewidth]{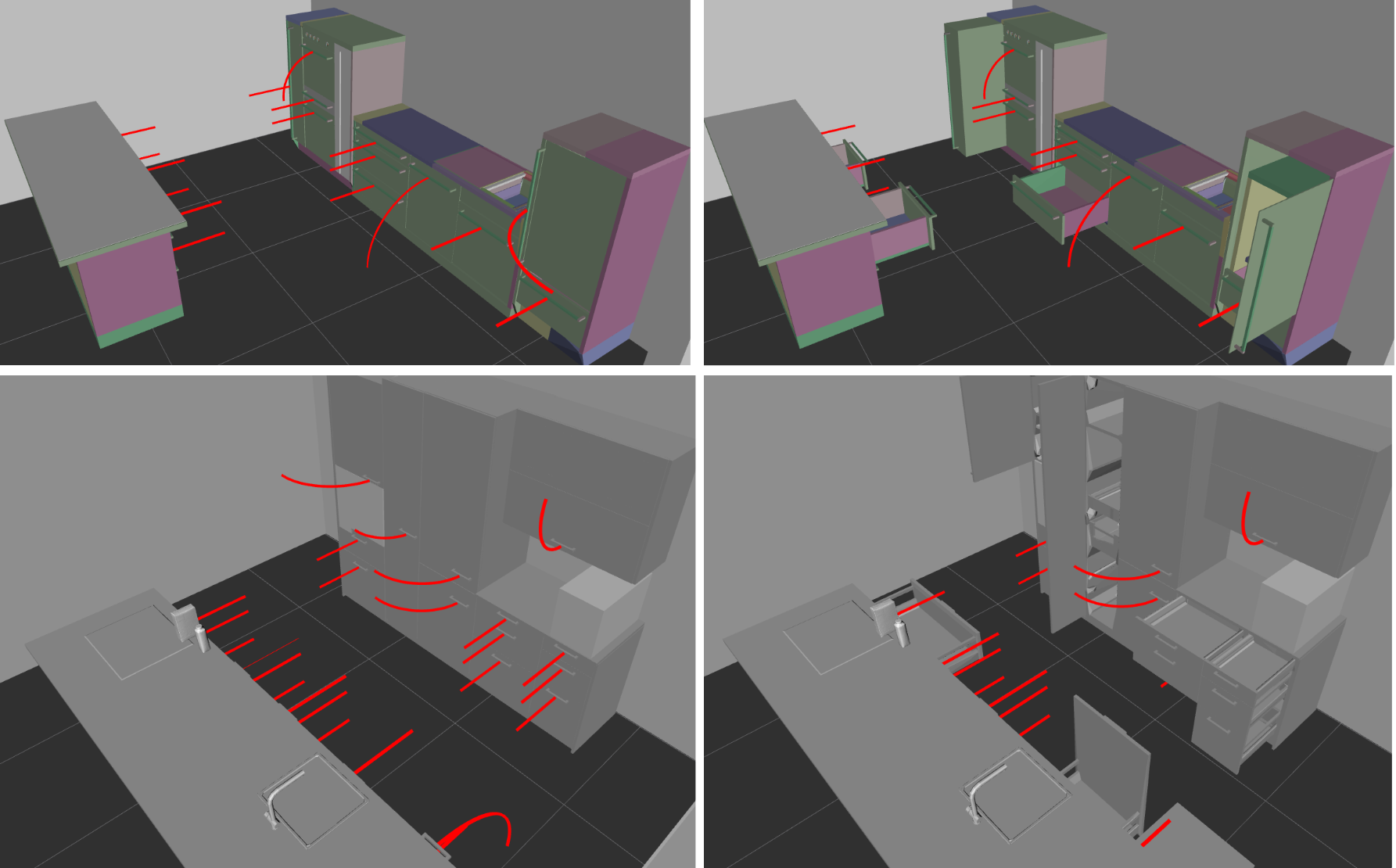}
  \caption{Motions $\tau$ (red lines) generated by evaluating  $\mathit{SatisfiesRequest}(\text{open()}, G_{\text{final}}) \wedge \mathit{Causes}(\tau, G_{\text{final}}, \Phi, I_{\Phi})$ for the kitchen B (top) and kitchen A (bottom) in distinct states. The left sides shows the results when all containers are initially closed. The right side shows the results when some containers are randomly selected to be open. No motion exists for containers that are already open to become open.}
  \label{fig:causesResults}
\end{figure}

The Law can be used to generate body motions from a task request by evaluating the predicates in order.
We exercise this here by evaluating the task request $\textit{open}(\text{door1})$.

From the predicate definitions in the previous section we know that this will first infer all final states in the SDT that satisfy the request, which is $G_{\text{final}} := p_{\text{door1}} \ge 0.9 p_{\text{door1}}^{\text{upper-limit}}$.
Next, we evaluate $\mathit{Causes}()$. If the state is already satisfied we return false, because the final state already exists and cannot be caused right now. Otherwise, we generate a trajectory of the joint value that causes the goal state to be true.

This can be seen in Fig.~\ref{fig:causesResults} where the same kitchen is shown in different states of opened drawers. In both cases the same task specification was evaluated. It can be seen that querying for the same task specification can yield different results depending of the current state of the environment. For our current example \textit{door1} is the rightmost in the top left image. We can see the handle motion for the case that the door is not already opened as the red line.

This motion now acts as the parameterization for the motion planner wrapped by the last predicate.
It can then be evaluated for any robot $R$ of interest to decide whether that robot can achieve the task request in the current environment.

This is, of course, a conservative estimation of the success of the manipulation action, as a possible failure during motion planning could be circumvented by using a different motion planner, or our \texttt{Causes()} evaluation could be inaccurate because the door was glued shut in reality.  
Nonetheless, we can now guarantee the success of the manipulation action with respect to the assumptions we made.
Fig.~\ref{fig:va} shows snapshots of the PR2 in a real kitchen opening a drawer after evaluating its motion planning as discussed in this use case.

\paragraph{Usage Mode 2: Verification of Given Motions.}
Instead of generating we can also verify if a motion causally and semantically satisfies a task request while being embodiment feasible.

First, assume there is an external motion generator such as a generative AI model, a recording of a human doing a task or any other method.
Second, flip the order of the predicate evaluation by starting to evaluate $\mathit{Causes}()$ with a fixed $\tau$ and a non fixed $G_{\text{final}}$.

The replay of the motion in the physics model will cause a final state that can then be compared to the expected final state of $\mathit{SatisfiesRequest}()$ for the task request.
Similarly, $\mathit{CanPerfom}()$ is evaluated to certify that the motion is embodiment feasible.
Choosing a different motion definition in the Law would allow to certify embodied motions for causality and semantic correctness.

\paragraph{Usage Mode 3: Learning from Motion Observations}
Replicating the previous usage mode but with a non fixed task request allows to infer which task request would cause a motion that was recorded from an observation, given a suitable request exist in the TEE class definition.
While this can be useful when a robotic agent tries to understand what a human is doing in the field of human robot interaction, which is out of the scope of this paper, the motion observation can also be used to extend the physics model.

Observing that a motion causes a final state that satisfies a known task request, but the $\mathit{Causes}()$ predicate cannot replicate that under the defined physics model is an indicator for a shortcoming in the physics model.
This can be intentional or it can be used as an opportunity to extend the model by learning from the observed data.

An example in $D_{\mathrm{artic}}$ can be a drawer that has a mechanism to open on its own after being pushed. 
This kind of articulation is not modelled in $\Phi_{\text{artic}}$ but it could be extracted from the observation that a pushing motion on the drawer causes it to open to create an updated physics model $\Phi_{\text{artic}}'$

\begin{figure}
  \centering
  \includegraphics[width=\linewidth]{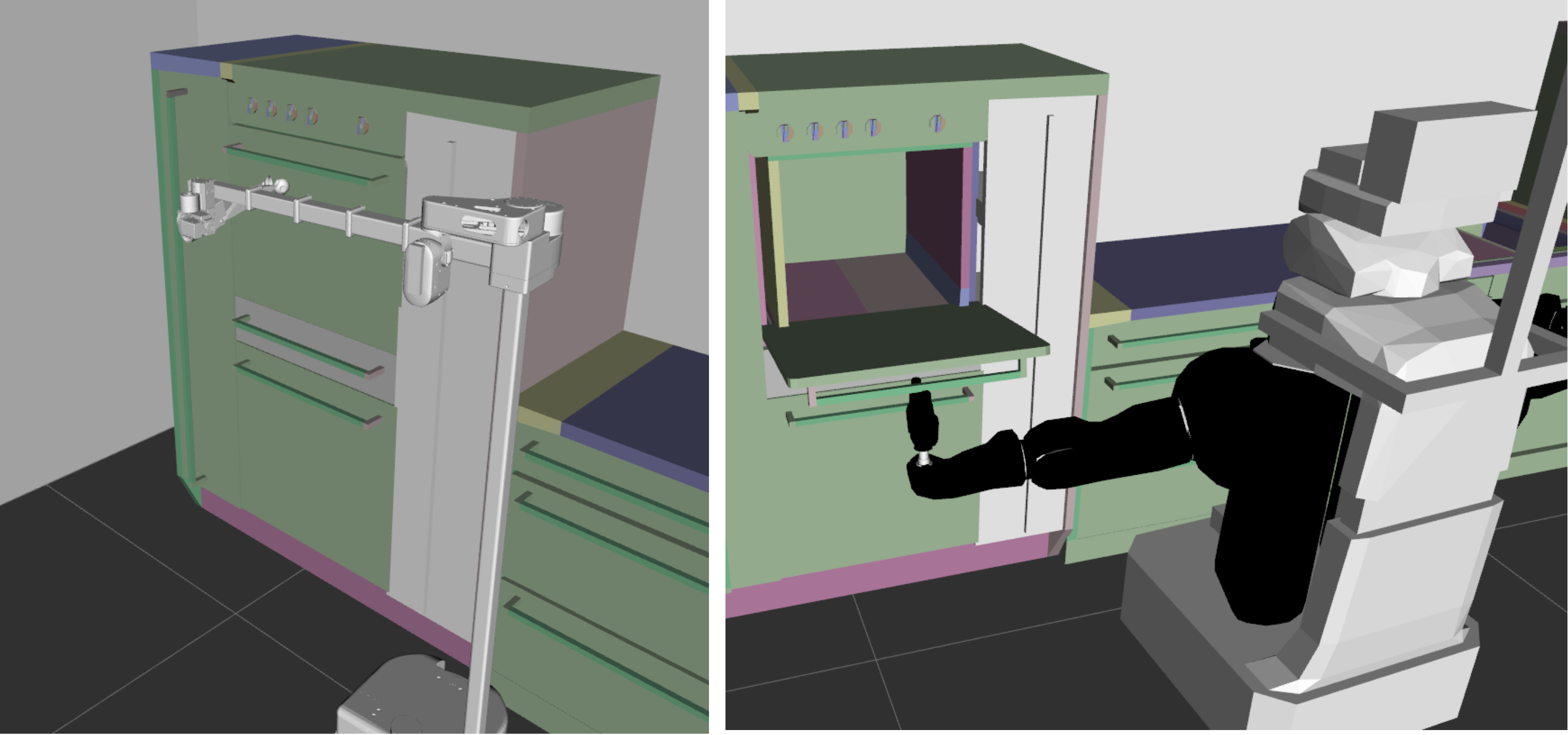}
  \caption{The left side shows a \textit{CanPerform}() failure of the Stretch robot for the task \textit{open}(oven door). Due to its limited reachability the Stretch is not able to perform the opening motion that starts with the gripper at the handle. The right side shows the PR2 being able to perform that task.}
  \label{fig:placeholder-failure}
\end{figure}
\paragraph{Usage Mode 4: Success Analysis Across Embodiments, Tasks and Environments.}
This can be realized with the law by selecting appropriate instances of the robot, environment or task class before evaluating the law.

For the domain $D_{\mathrm{artic}}$ we can, for example, ask: given a kitchen with all containers closed/opened what tasks can be performed; given a half opened door can this embodiment open/close it fully; can this container be opened when an obstacle is placed in front of it.

As a practical evaluation we use the Law implemented as described in the previous section and in accordance with usage mode 1 for the three robot embodiment's and two different kitchen environments, where every container is initially closed.
As a task request we use \textit{open()}, such that every container is considered. Kitchen A has 8 doors and 19 drawers, and Kitchen B has 3 doors and 14 drawers.

Table~\ref{tab:feasibility-matrix} shows which embodiment could achieve how many container opening tasks in which kitchen, while
Fig.~\ref{fig:placeholder-failure} shows a qualitative example of why one embodiment failed where another one succeeded. In this case the oven door opened downwards which is no problem for the PR2 since its arm has enough reach to be moved along the opening trajectory. However, the Stretch robot has an arm which can only be extended along one axis and rotate the gripper this makes it impossible for him to follow the downward opening trajectory of the door. This shows that the law can effectively verify if a specific embodiment is able to fulfill a given task request. 

This analysis enables structured robot and environment co-design. For a given kitchen, it can be identified which robot provides the broadest task coverage, or which environment modifications would make all tasks achievable by a specific robot.

\begin{table}
  \centering
  \small
  \caption{Results of checking which embodiment can achieve how many container openings in each environment}
  \label{tab:feasibility-matrix}
  \begin{tabular}{@{}l ccc@{}}
    \hline
     & PR2 & TIAGo++ & Stretch \\
    \hline
    \texttt{Kitchen A} & 22/27 & 22/27 & 19/27 \\
    \texttt{Kitchen B} & 16/17 & 16/17 & 15/17 \\
    
    \hline
  \end{tabular}
\end{table}

\section{Related Work}
\label{sec:rel}
We position the Law of Task-Achieving Body Motion at the intersection of task and motion planning, physics-based reasoning and digital twins, learning-based control, safety and verification, and recent neuro-symbolic and LLM-based planning. We structure this discussion by which component of the Law they primarily support: semantics, causality, or embodiment.

Grounding symbolic requests in geometric action is central to Task and Motion Planning (TAMP). Classical TAMP approaches~\cite{tamp_survey,garrett_tamp} use formal logics to sequence discrete actions and check feasibility via geometric predicates, while Logic-Geometric Programming (LGP)~\cite{toussaint_lgp} unifies these in a continuous--discrete optimization over modes and trajectories. These methods directly support \textit{SatisfiesRequest} and a geometric form of \textit{CanPerform} for rigid-body tasks in static environments but generally rely on kinematic abstractions with simplified contacts and therefore only partially address \textit{Causes} when richer physics such as frictional contact, fluid flow, or deformation matter.

Causal effects of actions have been modeled with learned and analytical physics engines for object interactions~\cite{battaglia_intuitive_physics,agrawal_learning_physics}. The industrial notion of a Digital Twin~\cite{glaessgen_digital_twin} emphasizes high-fidelity replication of physical systems for monitoring and prediction. Differentiable physics engines~\cite{degrave_diffphys} support gradient-based optimization through simulation and scene-graph frameworks~\cite{rosinol_kimera} provide structured 3D maps with semantics. These approaches supply strong machinery for \textit{Causes} by predicting what happens when actions are applied, but they typically do not define when an outcome is correct for an open-ended manipulation verb or explicitly mark out-of-scope regimes.

Large-scale robot learning is a dominant paradigm in manipulation. Domain randomization~\cite{tobin_domain_randomization,openai_rubik} trains policies across wide parameter ranges to obtain robustness to appearance and dynamics variation, and end-to-end visuomotor policies~\cite{rt1,octo} map pixels to actions with strong performance on diverse tasks. These methods primarily address \textit{CanPerform} and local robustness but lack an explicit verification interface, making it difficult to disentangle failures due to task semantics, physics mismatch, or embodiment limitations.
Safety focused work~\cite{safety_gym,safe_shielding} introduces failure modes and shielding, typically via state or action constraints under fixed dynamics and specification languages rather than task-level semantics under scoped physics.

Neuro-Symbolic AI~\cite{neurosymbolic_survey} aims to combine logical structure with the generalization abilities of deep learning, and recent systems use Large Language Models as high-level planners that translate natural language into code, API calls, or waypoints for robots~\cite{saycan,voxposer}. These approaches provide rich priors over everyday tasks but can hallucinate physically impossible or unsafe behaviors and usually lack a grounded, task-level notion of correctness. Formal verification and safety in robotics have been addressed through control-theoretic methods, shielded reinforcement learning, and runtime monitoring~\cite{safety_gym,safe_shielding}, typically over continuous-state models and low-level safety constraints.

Our work complements these efforts by providing a task-level axiom schema that combines semantics, causality, and embodiment within TEE classes. Rather than introducing a new planner or controller, the Law wraps existing TAMP, LGP, physics-based, or learning-based systems as an external certifier, with SDTs acting as semantic digital twins linking task descriptions, physical parameters, and embodiment models. As shown in Section~\ref{sec:evaluation}, this yields typed failure explanations (semantic, causal, embodiment) and environment-level analyses that are difficult to obtain from monolithic end-to-end approaches alone.

\section{Conclusion}
\label{sec:conc}
We treated open-world manipulation as a rigorous engineering problem and introduced the Law of Task-Achieving Body Motion, a universal axiom schema that connects task requests to physical execution through three verifiable predicates: \textit{SatisfiesRequest}, \textit{Causes}, and \textit{CanPerform}, defined over Task–Embodiment–Environment (TEE) classes, Semantic Digital Twins (SDTs), and a factored physics model. The Law acts as an external certifier that wraps arbitrary motion generators and methods for causal and semantic reasoning.

We operationalized this axiomatization with Semantic Digital Twins and TEE classes that make model assumptions and validity intervals on scoped physics models explicit.
This conservative interval scoping yields tractable subproblems for verifying the success of manipulation actions.
We instantiated the Law for articulated container manipulation in kitchen environments using three distinct robot embodiments to demonstrate how the framework supports motion generation, verification, and feasibility analysis across different robots and environments.

Future work includes extending the library of TEE classes to more realistic and complex physics domains such as deformable objects, granular media, and contact-rich assembly, and strengthening formal guarantees of the Law by reporting on the results. A particularly promising direction is integrating the Law with robotic foundation models and LLM-based planning systems~\cite{saycan,voxposer}, to have generative models propose tasks and motions while the Law provides structured verification. 
The three predicates of the Law should be able to detect semantic hallucinations, physically impossible proposals, and embodiment violations, to enable verify-repair loops. This would move toward a hybrid architecture combining GenAI's open-world language grounding with the Law's task-level correctness guarantees for explicit domains, advancing toward a broader algorithm theory of body motion for everyday manipulation.

\bibliographystyle{named}
\bibliography{ijcai26}

@article{garrett_tamp,
  author  = {Caelan Reed Garrett and Rohan Chitnis and Rachel Holladay and Beomjoon Kim and Tom Silver and Leslie Pack Kaelbling and Tom{\'a}s Lozano-P{\'e}rez},
  title   = {Integrated Task and Motion Planning},
  journal = {Annual Review of Control, Robotics, and Autonomous Systems},
  year    = {2021},
  volume  = {4},
  pages   = {265--293},
  doi     = {10.1146/annurev-control-091420-084139},
  url     = {https://doi.org/10.1146/annurev-control-091420-084139}
}

@article{tamp_survey,
  author  = {Zhigen Zhao and Shuo Cheng and Yan Ding and Ziyi Zhou and Shiqi Zhang and Danfei Xu and Ye Zhao},
  title   = {A Survey of Optimization-based Task and Motion Planning: From Classical To Learning Approaches},
  journal = {IEEE/ASME Transactions on Mechatronics},
  year    = {2024},
  doi     = {10.1109/TMECH.2024.3452509},
  url     = {https://arxiv.org/abs/2404.02817}
}

@inproceedings{toussaint_lgp,
  author    = {Toussaint, Marc},
  title     = {Logic-geometric programming: an optimization-based approach to combined task and motion planning},
  year      = {2015},
  isbn      = {9781577357384},
  publisher = {AAAI Press},
  booktitle = {Proceedings of the 24th International Conference on Artificial Intelligence},
  pages     = {1930–1936},
  numpages  = {7},
  location  = {Buenos Aires, Argentina},
  series    = {IJCAI'15}
}

@article{battaglia_intuitive_physics,
  author  = {Peter W. Battaglia  and Jessica B. Hamrick  and Joshua B. Tenenbaum },
  title   = {Simulation as an engine of physical scene understanding},
  journal = {Proceedings of the National Academy of Sciences},
  volume  = {110},
  number  = {45},
  pages   = {18327-18332},
  year    = {2013},
  doi     = {10.1073/pnas.1306572110},
  url     = {https://www.pnas.org/doi/abs/10.1073/pnas.1306572110},
  eprint  = {https://www.pnas.org/doi/pdf/10.1073/pnas.1306572110}
}

@misc{agrawal_learning_physics,
  title         = {Learning to Poke by Poking: Experiential Learning of Intuitive Physics},
  author        = {Pulkit Agrawal and Ashvin Nair and Pieter Abbeel and Jitendra Malik and Sergey Levine},
  year          = {2017},
  eprint        = {1606.07419},
  archiveprefix = {arXiv},
  primaryclass  = {cs.CV},
  url           = {https://arxiv.org/abs/1606.07419}
}

@inproceedings{glaessgen_digital_twin,
  title  = {The Digital Twin Paradigm for Future NASA and U.S. Air Force Vehicles},
  author = {Edward H. Glaessgen and Doane Stargel},
  year   = {2012},
  url    = {https://api.semanticscholar.org/CorpusID:110572580}
}

@misc{degrave_diffphys,
  title         = {A Differentiable Physics Engine for Deep Learning in Robotics},
  author        = {Jonas Degrave and Michiel Hermans and Joni Dambre and Francis wyffels},
  year          = {2018},
  eprint        = {1611.01652},
  archiveprefix = {arXiv},
  primaryclass  = {cs.NE},
  url           = {https://arxiv.org/abs/1611.01652}
}

@article{rosinol_kimera,
  author  = {Antoni Rosinol and Andrew Violette and Marcus Abate and Nathan Hughes and Yun Chang and Jingnan Shi and Arjun Gupta and Luca Carlone},
  title   = {Kimera: From SLAM to spatial perception with 3D dynamic scene graphs},
  journal = {The International Journal of Robotics Research},
  volume  = {40},
  number  = {12-14},
  pages   = {1510-1546},
  year    = {2021},
  doi     = {10.1177/02783649211056674},
  url     = { 
             
             https://doi.org/10.1177/02783649211056674
             },
  eprint  = { 
             https://doi.org/10.1177/02783649211056674
             }
}

@inproceedings{tobin_domain_randomization,
  author    = {Tobin, Josh and Fong, Rachel and Ray, Alex and Schneider, Jonas and Zaremba, Wojciech and Abbeel, Pieter},
  title     = {Domain randomization for transferring deep neural networks from simulation to the real world},
  year      = {2017},
  publisher = {IEEE Press},
  url       = {https://doi.org/10.1109/IROS.2017.8202133},
  doi       = {10.1109/IROS.2017.8202133},
  booktitle = {2017 IEEE/RSJ International Conference on Intelligent Robots and Systems (IROS)},
  pages     = {23–30},
  numpages  = {8},
  location  = {Vancouver, BC, Canada}
}

@misc{openai_rubik,
  title         = {Solving Rubik's Cube with a Robot Hand},
  author        = {OpenAI and Ilge Akkaya and Marcin Andrychowicz and Maciek Chociej and Mateusz Litwin and Bob McGrew and Arthur Petron and Alex Paino and Matthias Plappert and Glenn Powell and Raphael Ribas and Jonas Schneider and Nikolas Tezak and Jerry Tworek and Peter Welinder and Lilian Weng and Qiming Yuan and Wojciech Zaremba and Lei Zhang},
  year          = {2019},
  eprint        = {1910.07113},
  archiveprefix = {arXiv},
  primaryclass  = {cs.LG},
  url           = {https://arxiv.org/abs/1910.07113}
}

@misc{rt1,
  title         = {RT-1: Robotics Transformer for Real-World Control at Scale},
  author        = {Anthony Brohan and Noah Brown and Justice Carbajal and Yevgen Chebotar and Joseph Dabis and Chelsea Finn and Keerthana Gopalakrishnan and Karol Hausman and Alex Herzog and Jasmine Hsu and Julian Ibarz and Brian Ichter and Alex Irpan and Tomas Jackson and Sally Jesmonth and Nikhil J Joshi and Ryan Julian and Dmitry Kalashnikov and Yuheng Kuang and Isabel Leal and Kuang-Huei Lee and Sergey Levine and Yao Lu and Utsav Malla and Deeksha Manjunath and Igor Mordatch and Ofir Nachum and Carolina Parada and Jodilyn Peralta and Emily Perez and Karl Pertsch and Jornell Quiambao and Kanishka Rao and Michael Ryoo and Grecia Salazar and Pannag Sanketi and Kevin Sayed and Jaspiar Singh and Sumedh Sontakke and Austin Stone and Clayton Tan and Huong Tran and Vincent Vanhoucke and Steve Vega and Quan Vuong and Fei Xia and Ted Xiao and Peng Xu and Sichun Xu and Tianhe Yu and Brianna Zitkovich},
  year          = {2023},
  eprint        = {2212.06817},
  archiveprefix = {arXiv},
  primaryclass  = {cs.RO},
  url           = {https://arxiv.org/abs/2212.06817}
}

@misc{octo,
  title         = {Octo: An Open-Source Generalist Robot Policy},
  author        = {Octo Model Team and Dibya Ghosh and Homer Walke and Karl Pertsch and Kevin Black and Oier Mees and Sudeep Dasari and Joey Hejna and Tobias Kreiman and Charles Xu and Jianlan Luo and You Liang Tan and Lawrence Yunliang Chen and Pannag Sanketi and Quan Vuong and Ted Xiao and Dorsa Sadigh and Chelsea Finn and Sergey Levine},
  year          = {2024},
  eprint        = {2405.12213},
  archiveprefix = {arXiv},
  primaryclass  = {cs.RO},
  url           = {https://arxiv.org/abs/2405.12213}
}

@article{safety_gym,
  author = {Ray, Alex and Achiam, Joshua and Amodei, Dario},
  title  = {{Benchmarking Safe Exploration in Deep Reinforcement Learning}},
  year   = {2019}
}

@misc{safe_shielding,
  title         = {Provably Safe Deep Reinforcement Learning for Robotic Manipulation in Human Environments},
  author        = {Jakob Thumm and Matthias Althoff},
  year          = {2022},
  eprint        = {2205.06311},
  archiveprefix = {arXiv},
  primaryclass  = {cs.RO},
  url           = {https://arxiv.org/abs/2205.06311}
}

@misc{neurosymbolic_survey,
  title         = {Towards Cognitive AI Systems: a Survey and Prospective on Neuro-Symbolic AI},
  author        = {Zishen Wan and Che-Kai Liu and Hanchen Yang and Chaojian Li and Haoran You and Yonggan Fu and Cheng Wan and Tushar Krishna and Yingyan Lin and Arijit Raychowdhury},
  year          = {2024},
  eprint        = {2401.01040},
  archiveprefix = {arXiv},
  primaryclass  = {cs.AI},
  url           = {https://arxiv.org/abs/2401.01040}
}

@misc{saycan,
  title         = {Do As I Can, Not As I Say: Grounding Language in Robotic Affordances},
  author        = {Michael Ahn and Anthony Brohan and Noah Brown and Yevgen Chebotar and Omar Cortes and Byron David and Chelsea Finn and Chuyuan Fu and Keerthana Gopalakrishnan and Karol Hausman and Alex Herzog and Daniel Ho and Jasmine Hsu and Julian Ibarz and Brian Ichter and Alex Irpan and Eric Jang and Rosario Jauregui Ruano and Kyle Jeffrey and Sally Jesmonth and Nikhil J Joshi and Ryan Julian and Dmitry Kalashnikov and Yuheng Kuang and Kuang-Huei Lee and Sergey Levine and Yao Lu and Linda Luu and Carolina Parada and Peter Pastor and Jornell Quiambao and Kanishka Rao and Jarek Rettinghouse and Diego Reyes and Pierre Sermanet and Nicolas Sievers and Clayton Tan and Alexander Toshev and Vincent Vanhoucke and Fei Xia and Ted Xiao and Peng Xu and Sichun Xu and Mengyuan Yan and Andy Zeng},
  year          = {2022},
  eprint        = {2204.01691},
  archiveprefix = {arXiv},
  primaryclass  = {cs.RO},
  url           = {https://arxiv.org/abs/2204.01691}
}

@misc{voxposer,
  title         = {VoxPoser: Composable 3D Value Maps for Robotic Manipulation with Language Models},
  author        = {Wenlong Huang and Chen Wang and Ruohan Zhang and Yunzhu Li and Jiajun Wu and Li Fei-Fei},
  year          = {2023},
  eprint        = {2307.05973},
  archiveprefix = {arXiv},
  primaryclass  = {cs.RO},
  url           = {https://arxiv.org/abs/2307.05973}
}

@article{knowrob,
  author  = {Moritz Tenorth and Michael Beetz},
  title   = {KnowRob: A knowledge processing infrastructure for cognition-enabled robots},
  journal = {The International Journal of Robotics Research},
  volume  = {32},
  number  = {5},
  pages   = {566-590},
  year    = {2013},
  doi     = {10.1177/0278364913481635},
  url     = { 
             
             https://doi.org/10.1177/0278364913481635
             
             
             
             },
  eprint  = { 
             
             https://doi.org/10.1177/0278364913481635
             
             
             
             }
}

@inproceedings{qsr_3d,
  author    = {Sharma, Akash and Dong, Wei and Kaess, Michael},
  title     = {Compositional and Scalable Object SLAM},
  year      = {2021},
  publisher = {IEEE Press},
  url       = {https://doi.org/10.1109/ICRA48506.2021.9561697},
  doi       = {10.1109/ICRA48506.2021.9561697},
  booktitle = {2021 IEEE International Conference on Robotics and Automation (ICRA)},
  pages     = {11626–11632},
  numpages  = {7},
  location  = {Xi'an, China}
}

@inproceedings{mujoco,
  author    = {Todorov, Emanuel and Erez, Tom and Tassa, Yuval},
  booktitle = {2012 IEEE/RSJ International Conference on Intelligent Robots and Systems},
  title     = {MuJoCo: A physics engine for model-based control},
  year      = {2012},
  volume    = {},
  number    = {},
  pages     = {5026-5033},
  doi       = {10.1109/IROS.2012.6386109}
}

@misc{drake,
  author = {Russ Tedrake and the Drake Development Team},
  title  = {Drake: Model-based design and verification for robotics},
  year   = 2019,
  url    = {https://drake.mit.edu}
}

@misc{pybullet,
  author       = {Erwin Coumans and Yunfei Bai},
  title        = {PyBullet, a Python module for physics simulation for games, robotics and machine learning},
  howpublished = {\url{http://pybullet.org}},
  year         = {2016--2021}
}

@inproceedings{tracik,
  author    = {Beeson, Patrick and Ames, Barrett},
  booktitle = {2015 IEEE-RAS 15th International Conference on Humanoid Robots (Humanoids)},
  title     = {TRAC-IK: An open-source library for improved solving of generic inverse kinematics},
  year      = {2015},
  volume    = {},
  number    = {},
  pages     = {928-935},
  doi       = {10.1109/HUMANOIDS.2015.7363472}
}

@article{moveit,
  title   = {Reducing the barrier to entry of complex robotic software: a moveit! case study},
  author  = {Coleman, David and Sucan, Ioan and Chitta, Sachin and Correll, Nikolaus},
  journal = {arXiv preprint arXiv:1404.3785},
  year    = {2014}
}

@inproceedings{whole_body_mpc,
  author    = {Dantec, Ewen and Naveau, Maximilien and Fernbach, Pierre and Villa, Nahuel and Saurel, Guilhem and Stasse, Olivier and Taix, Michel and Mansard, Nicolas},
  booktitle = {2022 IEEE-RAS 21st International Conference on Humanoid Robots (Humanoids)},
  title     = {Whole-Body Model Predictive Control for Biped Locomotion on a Torque-Controlled Humanoid Robot},
  year      = {2022},
  volume    = {},
  number    = {},
  pages     = {638-644},
  doi       = {10.1109/Humanoids53995.2022.10000129}
}

@inproceedings{fcl_collision,
  author    = {Pan, Jia and Chitta, Sachin and Manocha, Dinesh},
  booktitle = {2012 IEEE International Conference on Robotics and Automation},
  title     = {FCL: A general purpose library for collision and proximity queries},
  year      = {2012},
  volume    = {},
  number    = {},
  pages     = {3859-3866},
  doi       = {10.1109/ICRA.2012.6225337}
}

@InProceedings{2001_counterfactual,
author="Ferrario, Roberta",
editor="Akman, Varol
and Bouquet, Paolo
and Thomason, Richmond
and Young, Roger",
title="Counterfactual Reasoning",
booktitle="Modeling and Using Context",
year="2001",
publisher="Springer Berlin Heidelberg",
address="Berlin, Heidelberg",
pages="170--183",
isbn="978-3-540-44607-1"
}

@article{roefer2022kineverse,
	author={Röfer, Adrian and Bartels, Georg and Burgard, Wolfram and Valada, Abhinav and Beetz, Michael},
	journal={IEEE Robotics and Automation Letters},
	title={Kineverse: A Symbolic Articulation Model Framework for Model-Agnostic Mobile Manipulation},
	year={2022},
	volume={7},
	number={2},
	pages={3372-3379},
	doi={10.1109/LRA.2022.3146515}
}

@phdthesis{stelter25giskard,
	author = {Simon Stelter},
	title = {A Robot-Agnostic Kinematic Control Framework: Task Composition via Motion Statecharts and Linear Model Predictive Control},
	year = {2025},
	doi = {10.26092/elib/3743},	
}
\end{document}